\pdfoutput=1

\documentclass[11pt]{article}

\usepackage[preprint]{coling}

\usepackage{times}
\usepackage{latexsym}

\usepackage[T1]{fontenc}

\usepackage[utf8]{inputenc}

\usepackage{microtype}

\usepackage{inconsolata}

\usepackage{graphicx}

\usepackage{hyperref}
\usepackage{booktabs}
\usepackage{multirow}
\usepackage{diagbox}
\usepackage{tabularx}
\usepackage{arydshln}
\usepackage{enumitem}
\usepackage{CJKutf8}
\usepackage{subfigure}
\usepackage{amssymb}
\usepackage{amsmath}
\usepackage{amssymb}
\usepackage{multirow}
\usepackage{color, xcolor}
\usepackage{pifont}
\usepackage[normalem]{ulem}
\useunder{\uline}{\ul}{}

\title{Hit the Sweet Spot! Span-Level Ensemble for Large Language Models}

\author{
    Yangyifan Xu\textsuperscript{1,2}, \ 
    Jianghao Chen\textsuperscript{1,2},\ 
    Junhong Wu\textsuperscript{1,2}, \
    Jiajun Zhang\textsuperscript{1,2,3,4}\thanks{\ \ Corresponding Author}   \\
    \textsuperscript{1}School of Artificial Intelligence, University of Chinese Academy of Sciences\\
    \textsuperscript{2}Institute of Automation, Chinese Academy of Sciences\\
    \textsuperscript{3}Wuhan AI Research, 
    \textsuperscript{4}Shanghai Artificial Intelligence Laboratory, Shanghai, China\\
    \texttt{\{xuyangyifan2021, chenjianghao2022, wujunhong2021\}@ia.ac.cn}, \texttt{jjzhang@nlpr.ia.ac.cn} \\
}

\begin{document}
\maketitle
\begin{abstract}
Ensembling various LLMs to unlock their complementary potential and leverage their individual strengths is highly valuable.
Previous studies typically focus on two main paradigms: sample-level and token-level ensembles.
Sample-level ensemble methods either select or blend fully generated outputs, which hinders dynamic correction and enhancement of outputs during the generation process. 
On the other hand, token-level ensemble methods enable real-time correction through fine-grained ensemble at each generation step. However, the information carried by an individual token is quite limited, leading to suboptimal decisions at each step.
To address these issues, we propose \textbf{\textsc{SweetSpan}}, a span-level ensemble method that effectively balances the need for real-time adjustments and the information required for accurate ensemble decisions. Our approach involves two key steps: First, we have each candidate model independently generate candidate spans based on the shared prefix. 
Second, we calculate perplexity scores to facilitate mutual evaluation among the candidate models and achieve robust span selection by filtering out unfaithful scores.
To comprehensively evaluate ensemble methods, we propose a new challenging setting (ensemble models with significant performance gaps) in addition to the standard setting (ensemble the best-performing models) to assess the performance of model ensembles in more realistic scenarios.
Experimental results in both standard and challenging settings across various language generation tasks demonstrate the effectiveness, robustness, and versatility of our approach compared with previous ensemble methods.
\end{abstract}

\section{Introduction}
Recently, large language models (LLMs) have rapidly developed, leading to the emergence of numerous diverse models~\citep{touvron2023llama,vicuna2023}.
These LLMs differ in datasets, architectures, and training methodologies, each exhibiting its own strengths and weaknesses~\citep{llm-blender-2023}. Therefore, ensembling these LLMs to unleash their complementary potential and leverage their individual strengths is highly valuable~\citep{llm-blender-2023,lu2023routing,shnitzer2023large}. Model ensembling has thus become a key focus and an active topic of current LLM research~\citep{lu2024merge}.

\begin{figure}[t]
    \centering
    \includegraphics[width=\hsize]{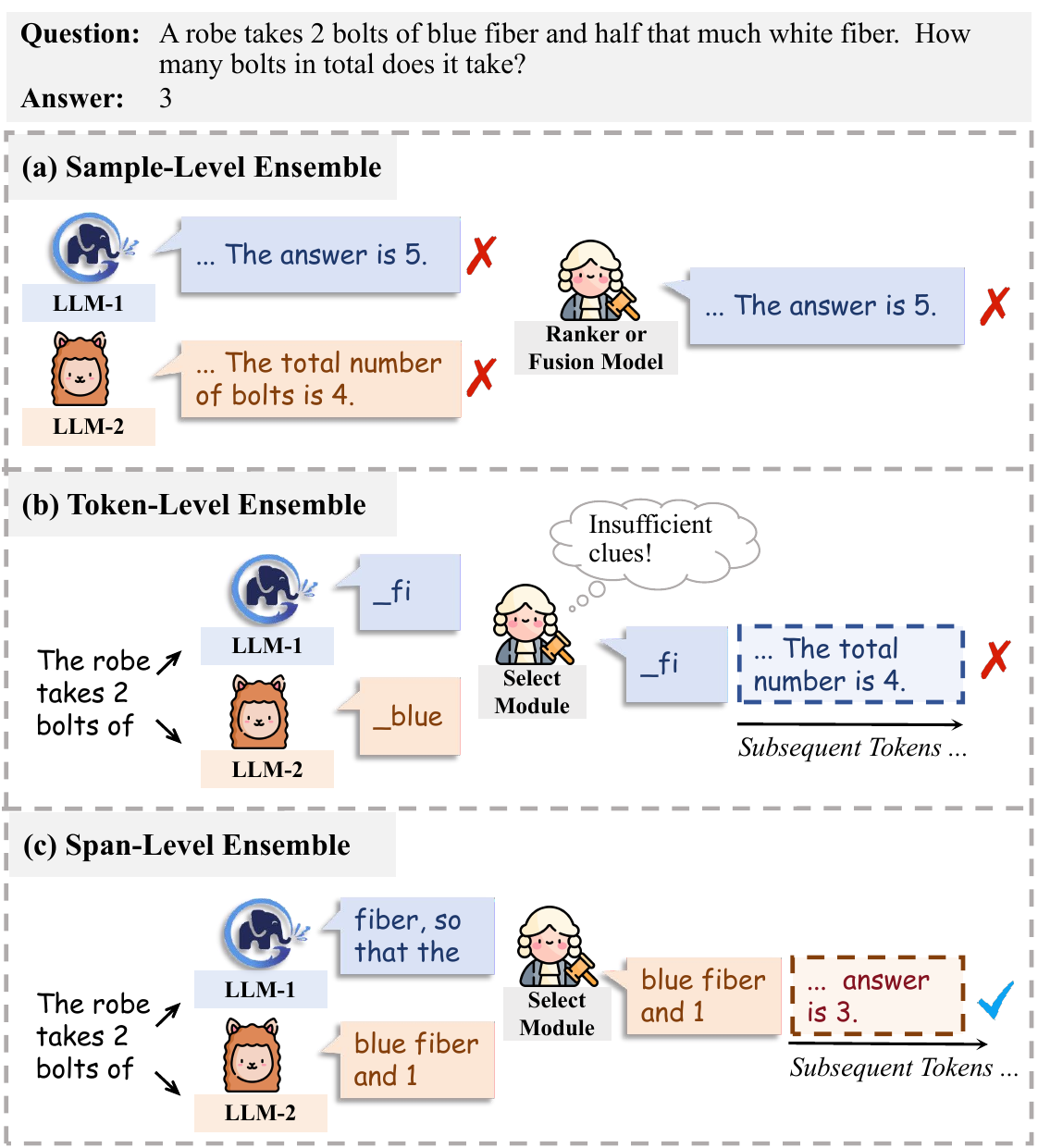}
    \caption{\textbf{Motivation of \textsc{SweetSpan}.} Sample-level ensemble methods struggle to produce a correct answer when all candidate outputs are flawed, while token-level ensemble methods make suboptimal choices at each generation step due to inadequate information. \textsc{SweetSpan} balances the flexibility needed for real-time adjustments and the information required for accurate ensemble decisions at each step.}
    \label{fig.motivation}
\end{figure}

Existing approaches can be classified into two categories:
1) Sample-level ensemble methods use an additional model to either select~\cite{lu2023routing,shnitzer2023large} or blend~\cite{llm-blender-2023} fully generated outputs.
As a result, these methods are unable to correct and enhance outputs during the generation process.
As illustrated in Figure~\ref{fig.motivation}(a), they struggle to produce a correct answer when all candidate outputs are flawed.
Moreover, the reliance on an additional model poses challenges for generalization to unseen data distributions.
2) Token-level ensemble methods achieve fine-grained ensemble at each generation step through output distribution alignment and thus enable real-time correction~\cite{xu2024bridging, huang2024enabling, yu2024breaking}. However, as shown in Figure~\ref{fig.motivation}(b), since tokens are smaller units than words and vary across models, the information they provide is often insufficient and inaccurate. This leads to suboptimal decisions at each step, hindering the achievement of globally optimal outputs.

\begin{figure*}[ht!]
    \centering
    \includegraphics[width=\textwidth]{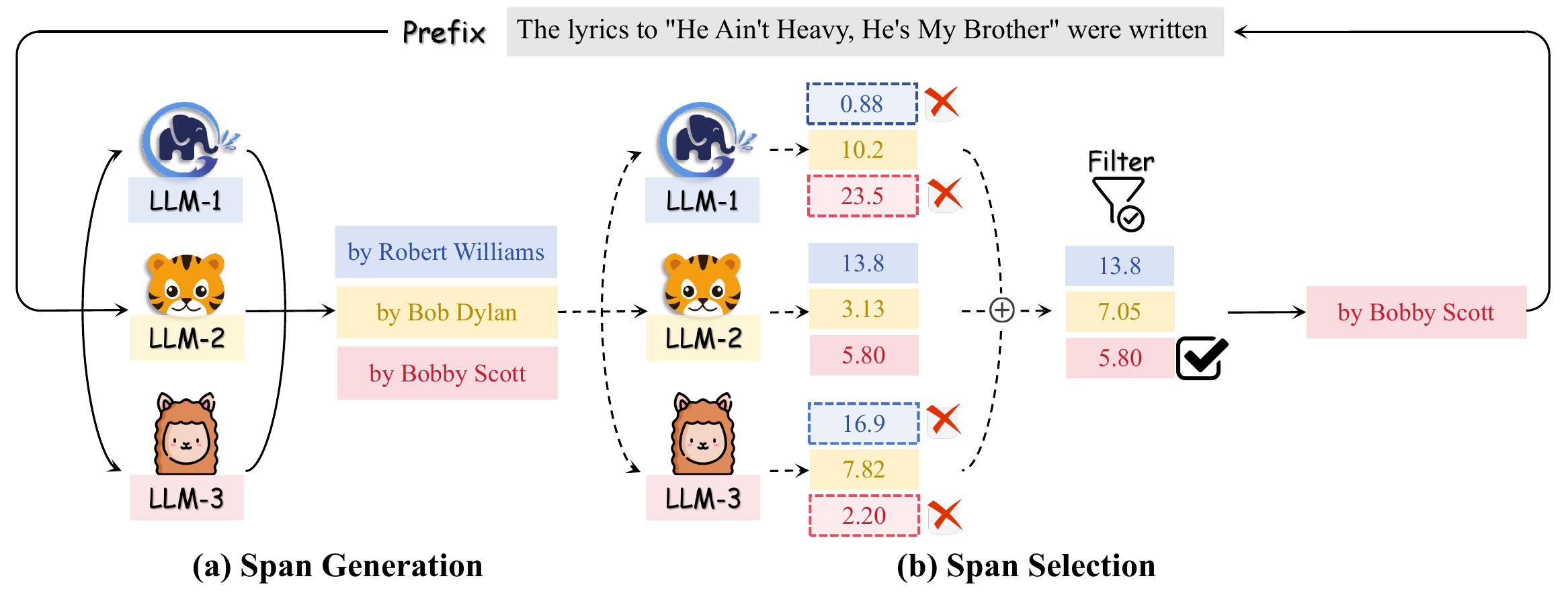}
    \caption{\textbf{The \textsc{SweetSpan} framework.} \textsc{SweetSpan} consists of two steps. (a) First, we have each candidate model generate a span based on the shared prefix. (b) Next, we facilitate mutual evaluation among the candidate models by calculating perplexity and achieve robust span selection by filtering out unfaithful scores.}
    \label{fig.framework}
\end{figure*}

To tackle the above issues, we propose a training-free span-level ensemble method named \textbf{\textsc{SweetSpan}}. Our method strikes a balance between the flexibility needed for real-time adjustments and the information required for accurate ensemble decisions at each step, hitting the sweet spot for the model ensemble.
Specifically, in each generation round, we first have each candidate model independently generate a span based on the shared prefix.
Subsequently, we facilitate mutual evaluation among the candidate models by calculating perplexity scores and filtering out unfaithful results to prevent underperforming models from skewing the evaluation. 
Finally, we choose the span with the lowest average perplexity as the ensemble result and attach it to the prefix for subsequent generations.

We evaluate our method on various language generation tasks, including commonsense reasoning, arithmetic reasoning, code generation, and machine translation.
We first ensemble the best-performing LLMs as a standard setting to assess the upper-bound performance of different ensemble methods. \textsc{SweetSpan} consistently achieves significant performance improvements across multiple tasks compared to previous methods, demonstrating its effectiveness and versatility.
We then ensemble LLMs with significant performance gaps as a challenging setting to evaluate the noise resistance of ensemble methods.
Unlike previous methods that collapse on most tasks, \textsc{SweetSpan}, aided by an effective filtering strategy, consistently delivers positive improvements, highlighting its robustness.

Briefly, our contributions can be summarized from the perspectives of effectiveness, robustness, and versatility:
\begin{itemize}
    \item We propose a span-level ensemble method that balances the flexibility needed for real-time adjustments and the information required for accurate ensemble decisions at each step. Experimental results demonstrate the effectiveness of our method.
    \item Beyond the standard setting, we introduce a new challenge setting that tests the noise resistance of ensemble methods by ensembling models with significant performance gaps. We observe that previous ensemble methods collapse on most tasks under this setting, while \textsc{SweetSpan} achieves stable positive improvements, highlighting its robustness.
    \item \textsc{SweetSpan} is not constrained by vocabulary, model architecture, or task, and can be directly applied to any LLM ensemble without additional parameter training, showcasing strong versatility.
\end{itemize}

\section{Our Method}
To achieve span-level ensemble for LLMs, we have to figure out two key issues. First, \textit{how to define spans?} Second, \textit{how to identify the optimal span?} We delve into these issues in the following subsections.

\subsection{How to define spans?}\label{sec:method1} 
As shown on the left side of Figure~\ref{fig.framework}, in each generation round, we first have each candidate model independently generate a text span, which has the following two characteristics: 1) Spans are composed of words rather than tokens. We ensure that spans do not cross word boundaries to prevent subsequent evaluations from being affected by different tokenizers across models. 2) All candidate spans have the same predefined length, such as 4. On the one hand, a longer span provides more information during the span selection phase, increasing the probability of making the optimal choice in the current round. On the other hand, a shorter span allows for more ensemble corrections over multiple rounds, reducing the probability of local errors. Therefore, an appropriate span length is crucial for balancing these competing demands and achieving optimal ensemble results.

\subsection{How to identify the optimal span?}\label{sec:method2}
We choose the simple and widely used perplexity as the evaluation metric for the span selection phase. 
Perplexity measures how well a language model fits a given span of text. A lower perplexity score indicates that the span aligns more closely with the internal knowledge of the language model, making it a reliable metric for span evaluation. 

Considering a set of $N$ candidate models (denoted as $\mathcal{M}$), we first have each candidate model ($m \in \mathcal{M}$) independently calculate the perplexity score for all candidate spans: 
\begin{equation*}
{\rm PPL}_{m}(\mathbf{s}_j) = \exp\left(-\frac{1}{|\mathbf{s}_j|} \sum_{t \in \mathbf{s}_j} \log p(t)\right),
\end{equation*}
where $\mathbf{s}_j$ is the candidate span generated by the $j$-th candidate model $m_j$.
Since all spans contain only complete words, vocabulary discrepancies do not affect the perplexity calculation.

Next, we filter the raw evaluation scores. 
Models lacking relevant knowledge of the current sample may assign unjustifiably perplexity scores, giving correct spans excessively high perplexity while assigning overly low perplexity to their own incorrect spans.
As shown in Figure~\ref{fig.framework}, under the current prefix, LLM-1, which fails to answer the question correctly, assigns an unjustifiably high perplexity score of 23.5 to the correct span "\textit{by Bobby Scott}", while giving its own incorrect span "\textit{by Robert Williams}" a low perplexity score of 0.88. As a result, without filtering, the correct span ranks below the incorrect ones, leading to an incorrect output.
To address this issue, we design an adaptive filtering strategy that identifies outliers based on the discrepancy in evaluation scores for each span. If the maximum score exceeds the minimum score by a factor of $\lambda$, we remove the highest and lowest scores for that span. 
As shown in Figure~\ref{fig.framework}, by filtering out the maximum value, we prevent LLM-1 from excessively increasing the perplexity for the correct span, thus avoiding an overly negative assessment. By filtering out the minimum value, we avoid LLM-1's overconfidence in the incorrect span.
Our proposed filtering method effectively removes unjustified outliers, thereby preventing them from skewing the evaluation.
\begin{align*}
\mathcal{R}_j &= \left\{ \underset{m \in \mathcal{M}}{\operatorname{argmax}} \, {\rm PPL}_{m}(\mathbf{s}_j), \, \underset{m \in \mathcal{M}}{\operatorname{argmin}} \, {\rm PPL}_{m}(\mathbf{s}_j) \right\} \\
\mathcal{C}_j &= 
\begin{cases} 
\mathcal{M} \setminus \mathcal{R}_j, & \text{if } \dfrac{\underset{m \in \mathcal{M}}\max {\rm PPL}_{m}(\mathbf{s}_j)}{\underset{m \in \mathcal{M}}\min {\rm PPL}_{m}(\mathbf{s}_j)} \textgreater \lambda  \\
\mathcal{M}, & \text{otherwise}
\end{cases}
\end{align*}

Finally, we aggregate the filtered perplexity scores and select the span with the lowest average perplexity as the ensemble result for the current generation round.
\begin{equation*}
\mathbf{s}^* = \underset{\mathbf{s}_j, 1 \leq j \leq N}{\arg\min}\frac{1}{|\mathcal{C}_{j}|}\sum_{m \in \mathcal{C}_{j}} {\rm PPL}_{m}(\mathbf{s}_j)
\end{equation*}

\section{Experimental Settings}

\subsection{Tasks and Datasets} 
To demonstrate the versatility of our method, we explore a wide range of language generation tasks:

\paragraph{Commonsense Reasoning via Natrual Question (NQ)~\citep{kwiatkowski2019natural}:}
Models are tested to answer questions originating from real queries issued to the Google search engine. 
The evaluation metric is Exact Match.

\paragraph{Arithmetic Reasoning via GSM8K~\citep{cobbe2021gsm8k}:}
Models are tasked to solve math problems that require multi-step reasoning. These problems are at the grade school level.
The evaluation metric is Accuracy.

\paragraph{Code Generation via MBPP~\citep{austin2021program}:}
Models are requested to generate code that solves basic Python programming problems. These problems are designed to be solvable by entry-level programmers, covering fundamentals and standard library functionality. The evaluation metric is Pass@1.

\paragraph{Machine Translation via Flores-101~\citep{flores101}:}
Machine translation is a traditional NLP task, which demonstrates the multilinguality of LLMs. We use the German (De)$\leftrightarrow$English (En) split and Romanian (Ro)$\leftrightarrow$English split for evaluation. The evaluation metric is BLEU~\citep{post-2018-call}.

\subsection{Candidate LLMs}
We choose seven open-source chat LLMs, each approximately 7B in size, as the candidate LLMs to form our ensemble: LLaMA2-7B-Chat~\citep{touvron2023llama}, ChatGLM2-6B~\citep{zeng2022glm}, Baichuan2-7B-Chat~\citep{baichuan2023baichuan2}, InternLM-7B-Chat~\citep{2023internlm}, TigerBot-7B-Chat-V3~\citep{chen2023tigerbot}, Vicuna-7B-V1.5~\citep{vicuna2023}, ChineseAlpaca2-7B~\citep{Chinese-LLaMA-Alpaca}.

Each model leverages large-scale, high-quality data to establish a strong knowledge base and is aligned by supervised instruction tuning, thereby achieving high performance on public benchmarks. Originating from distinct institutions, these models exhibit inherent diversity, which provides opportunities for effective ensemble.

\subsection{Baselines}
We compare \textsc{SweetSpan} with strong baseline methods, including both sample-level and token-level ensemble approaches.

\paragraph{PairRanker} \citet{llm-blender-2023} utilize a pairwise comparison to identify subtle distinctions between different candidate outputs.

\paragraph{LLM-Blender} \citet{llm-blender-2023} employ a 3B-parameter model fine-tuned on an instruction dataset to combine the ranking results from PairRanker and produce the final output.
\paragraph{EVA} \citet{xu2024bridging} utilizes overlapping tokens as anchors to map the output distributions of heterogeneous LLMs into a unified space, enabling a fine-grained token-level ensemble.

\subsection{Implement Details}
We utilize greedy decoding in all experiments since it generally produces higher-quality outputs~\citep{uzan2024greed}. 
Empirically, we set $\lambda = 10$, with a detailed analysis provided in Appendix~\ref{sec:lambda}.
For machine translation tasks, we employ a 4-shot in-context learning setting, while for other tasks, we perform zero-shot inference. Furthermore, we follow the required format for each chat model and use task-specific prompts, including the specific incorporation of a chain of thought prompt in the arithmetic reasoning task.

\section{Experimental Results and Analysis}
\subsection{Standard Setting: Ensembling Top-Performing Models}\label{sec:exp1}
\begin{figure*}[!t]
	\centering
	\small
        \subfigure[NQ]{
            \includegraphics[width=0.32\textwidth]{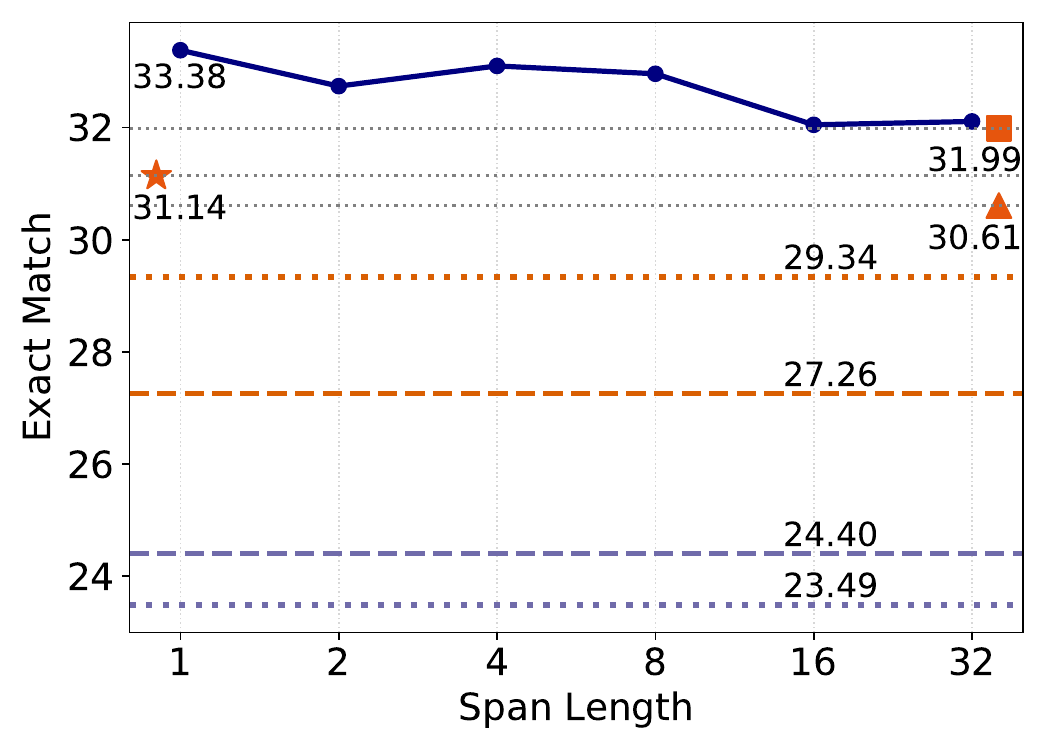} 
        }
        \subfigure[MBPP]{
            \includegraphics[width=0.32\textwidth]{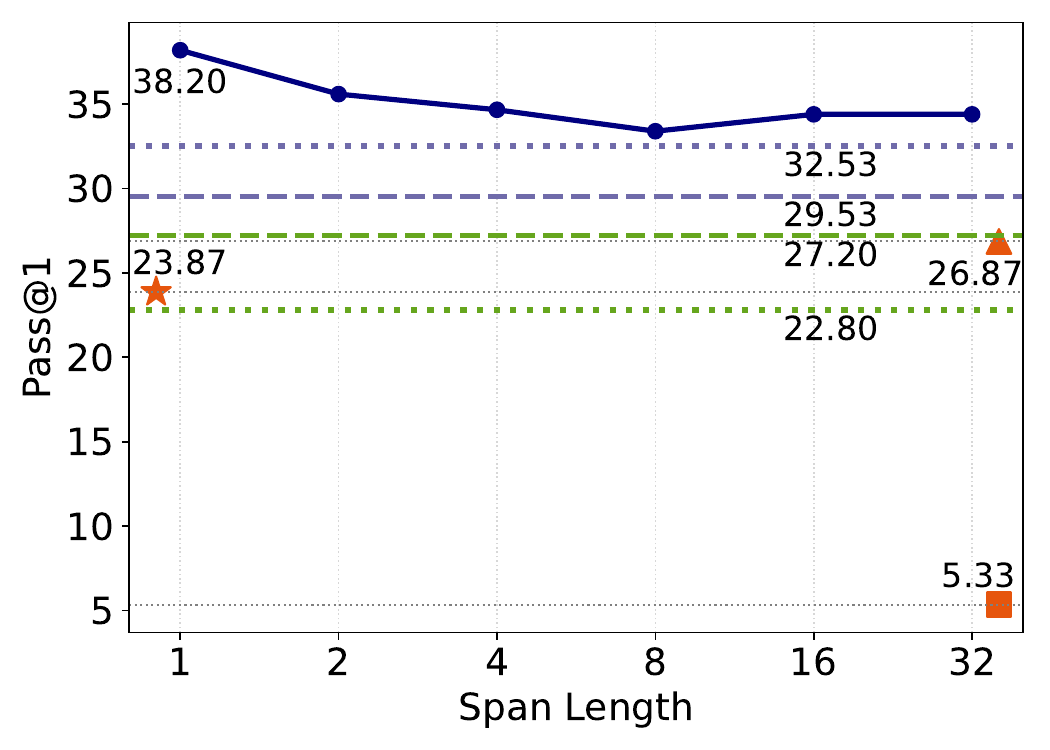}
        } 
        \subfigure[GSM8K]{
            \includegraphics[width=0.32\textwidth]{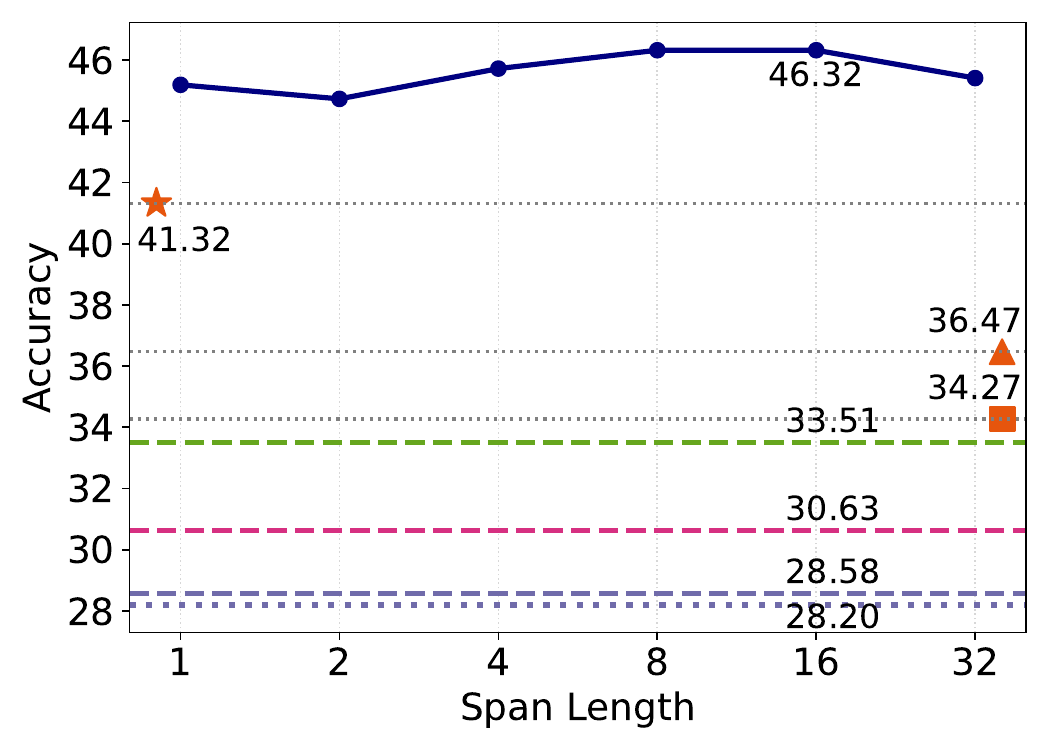} 
        } \\
        \subfigure[Flores En$\rightarrow$De]{
            \includegraphics[width=0.32\textwidth]{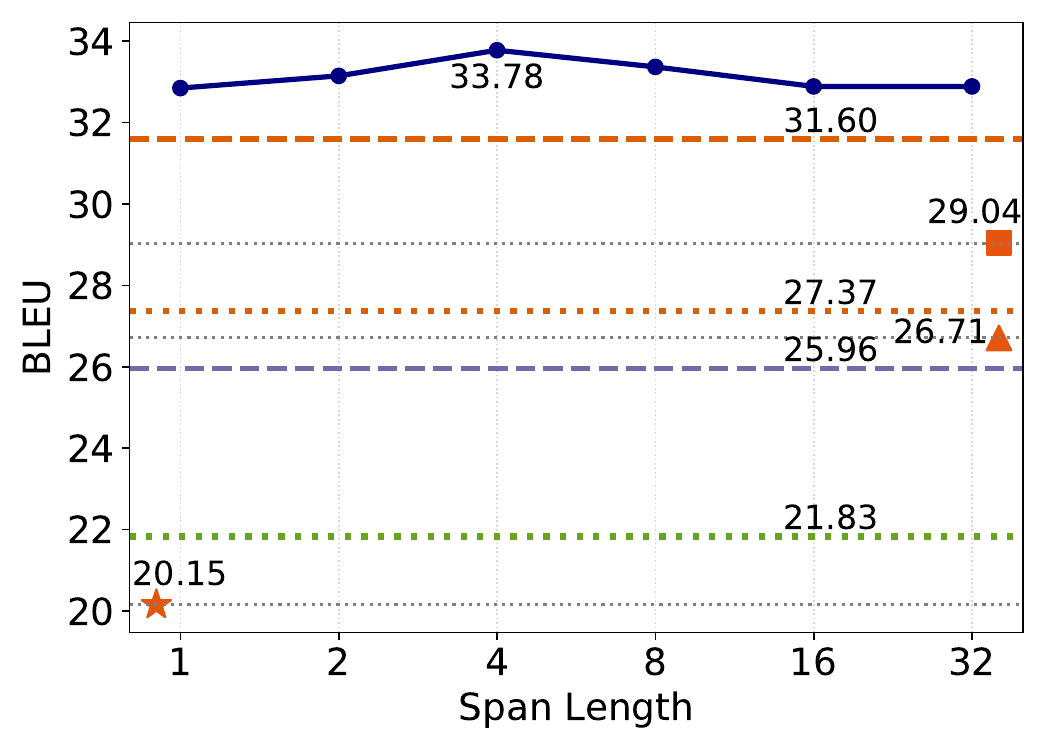} 
        } 
        \subfigure[Flores De$\rightarrow$En]{
            \includegraphics[width=0.32\textwidth]{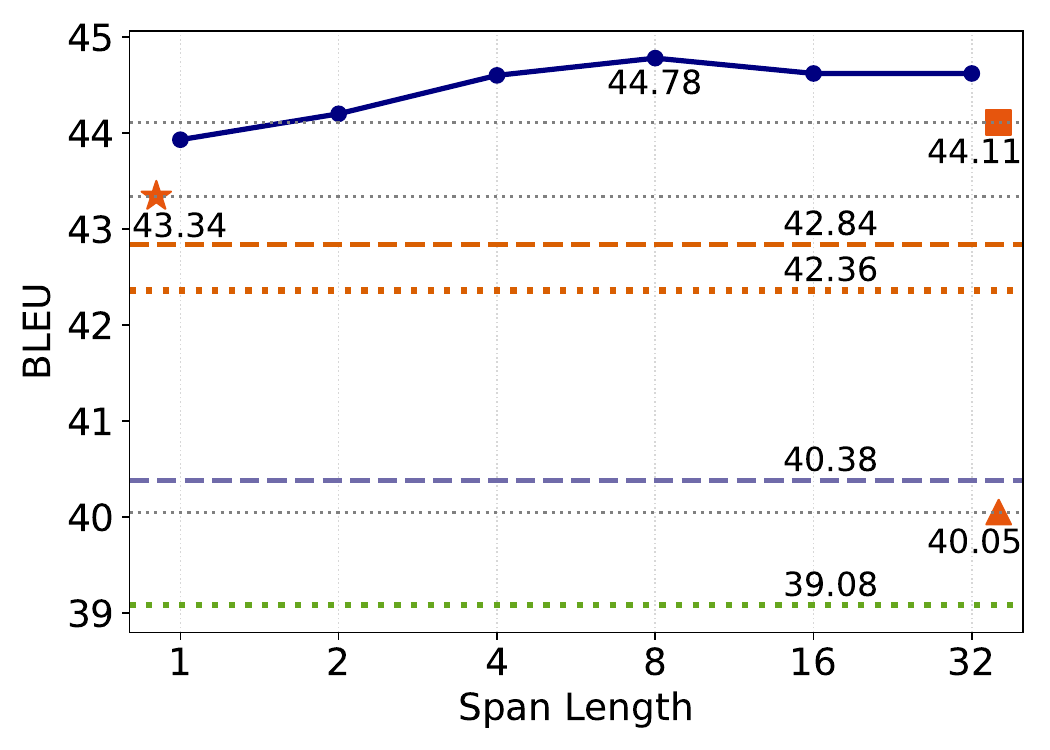}
        }
        \multirow{2}{*}[10ex]{ 
            \includegraphics[width=0.32\textwidth]{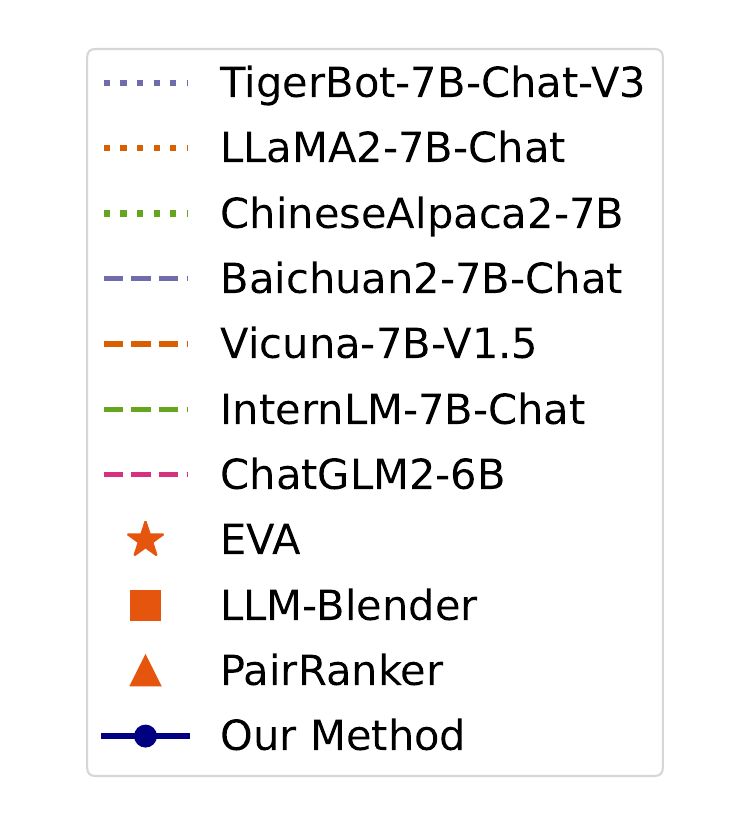}
        } \\
        \subfigure[Flores En$\rightarrow$Ro]{
            \includegraphics[width=0.32\textwidth]{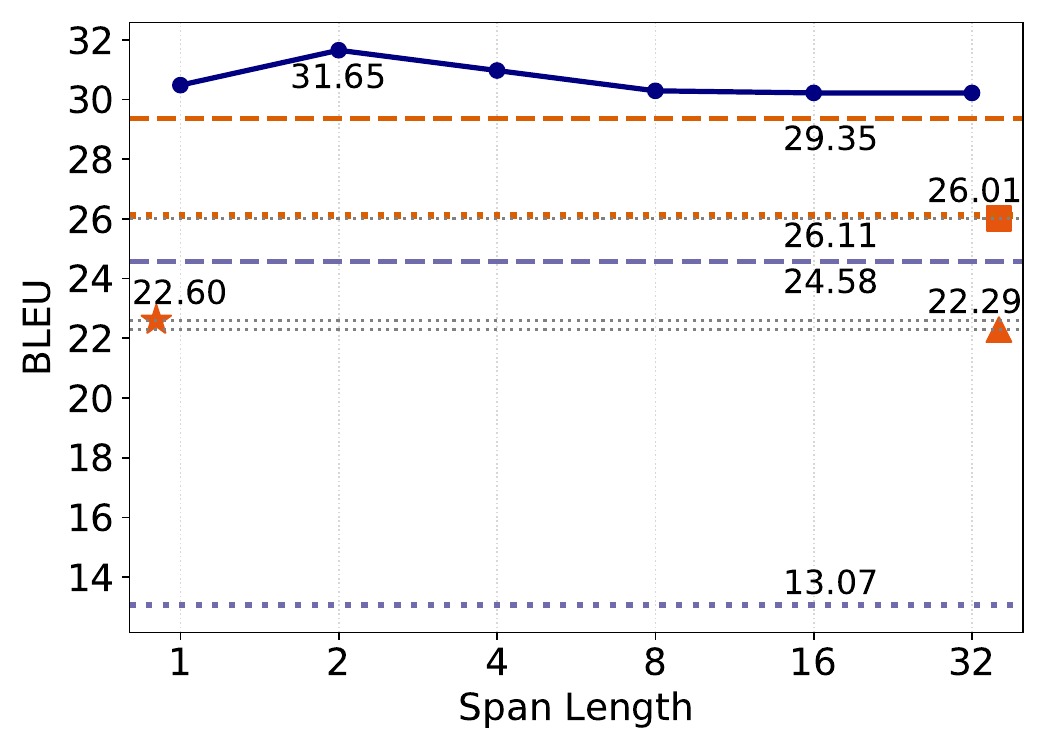} 
        } 
        \subfigure[Flores Ro$\rightarrow$En]{
            \includegraphics[width=0.32\textwidth]{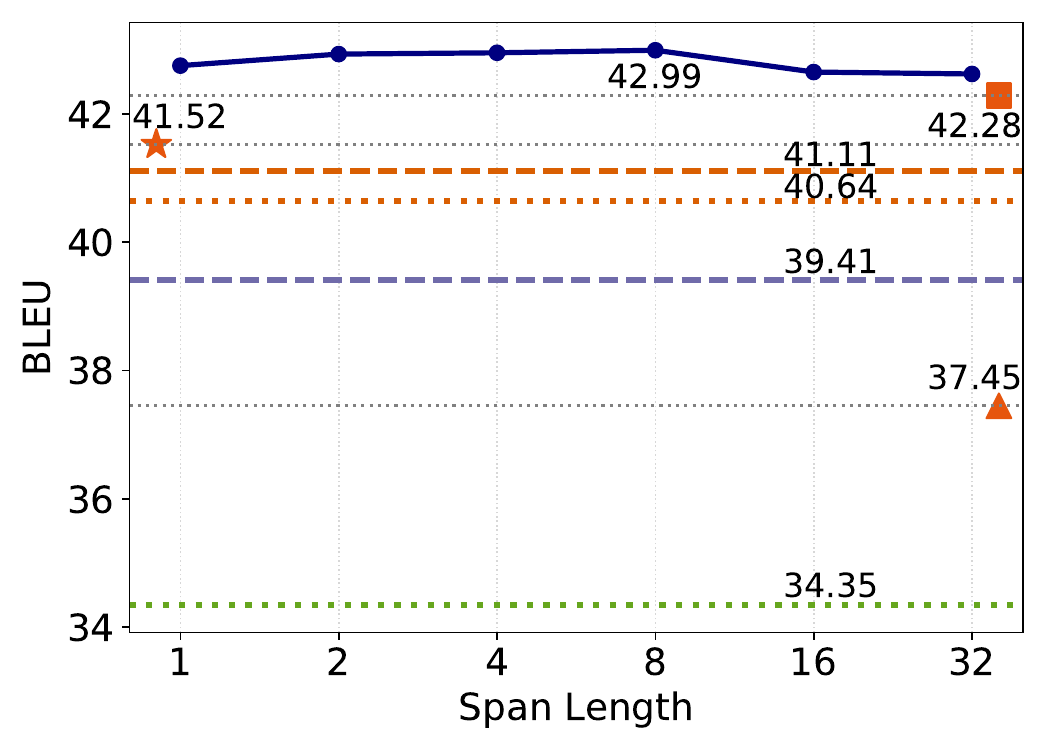}
        } 
        \hspace{0.32\textwidth} 
\caption{Main results on various language generation tasks under the standard setting.\protect\footnotemark}
\label{fig:main_result}
\end{figure*}

For each task, we select the top-performing four models out of seven for the ensemble, following a setup similar to previous work~\cite{xu2024bridging}.
We conduct experiments with span lengths of 1, 2, 4, 8, 16, and 32, and the trends in ensemble performance are shown in Figure~\ref{fig:main_result}.\footnote{Span length refers to the number of words within the span.}

\subsubsection{\textsc{SweetSpan} Demonstrates Superiority}
Our proposed \textsc{SweetSpan} consistently outperforms individual models and prior ensemble methods across all types of tasks at nearly all span lengths, demonstrating the effectiveness and cross-task versatility of our approach. 
Notably, in the GSM8K task, \textsc{SweetSpan} achieves a significant 12.21-point improvement compared with the best-performing individual model. 
We attribute this success to the appropriate granularity, which effectively balances the competing demands of information richness and flexibility.

\subsubsection{Guidelines for Determining Appropriate Span Length}
We observe that the optimal span length varies based on the specific characteristics of each task. The following guidelines provide recommendations for selecting an appropriate span length tailored to different tasks:

\paragraph{Deterministic tasks prefer shorter spans.}
Deterministic tasks are characterized by having precisely defined acceptable answers.
NQ, where the answer is a fixed phrase, and MBPP, where the answer is code, are examples of such tasks, both achieving optimal ensemble performance with a span length of 1.
As shown in Table~\ref{tab:case_study}, for the MBPP task, the strict syntax of code leads to a narrow range of correct answers, where a two-word span is already sufficient to confirm that the second span violates the syntax structure. In this case, longer spans do not provide additional useful information for the current decision and may even hinder timely correction of errors made during independent generation by candidate models.

In contrast, for the GSM8K task, which uses chain-of-thought prompting, intermediate reasoning steps can be expressed in various ways. A two-word span often doesn't provide enough information to make the best decision, and early incorrect choices can cause subsequent tokens to deviate from the intended meaning. In this case, a longer span is necessary to compare different reasoning paths. 
This aligns with the performance observed in GSM8K, where optimal ensemble performance is achieved with a span length of 16.

\footnotetext{Please refer to appendix~\ref{sec:main_result} for detailed information on the models used for each task and the experimental results presented in tabular form.}

\begin{table*}[]
\resizebox{\textwidth}{!}{
\begin{tabular}{@{}lll@{}}
\toprule[1.5pt]
\multicolumn{3}{c}{\textit{Code Generation Task: MBPP}}                                                                                                                         \\ \midrule 
Prefix           & \multicolumn{2}{l}{ def sort\_matrix(matrix):\textbackslash n\ \ \ \ n = len(matrix)\textbackslash n\ \ \ \ for}    \\ \cmidrule(l){2-3} 
2-word-candidate spans   &  i in &| \ \ \textbackslash n i \ \ \\ \cmidrule(l){2-3} 
8-word-candidate spans &    i in range(n):\textbackslash n\ \ \ \ for j in & | \ \ \textbackslash n i in range(n):\textbackslash n\ \ \ \ for j  \ \ \\ \midrule
\multicolumn{3}{c}{\textit{Arithmetic Reasoning Task: GSM8K}}                                                                                                                                \\ \midrule
Prefix           &   \multicolumn{2}{l}{1. Janet's ducks lay 16 eggs per day. }    \\ \cmidrule(l){2-3} 
2-word-candidate spans   &  She eats & | \ \  She bakes \ \  \\ \cmidrule(l){2-3} 
8-word-candidate spans &  She eats three for breakfast every morning so & | \ \  She bakes muffins with 4 eggs so she  \ \  \\  \bottomrule[1.5pt]
\end{tabular}}
\caption{\label{tab:case_study} 
Examples of spans with different lengths in GSM8K and MBPP tasks.}
\end{table*}

\paragraph{An intermediate span length is effective across all tasks.}
As illustrated in Figure~\ref{fig:main_result}, 
although different tasks exhibit varying trends, an intermediate span length, such as 4, consistently delivers significant improvements over candidate models. The intermediate span length effectively balances the need for real-time adjustments and the information required for accurate ensemble decisions at each ensemble step. We advise adopting a span length of 4 when applying our method to unknown tasks.

\subsubsection{\textsc{SweetSpan} Avoid the Inherent Limitations of Token-Level and Sample-Level Ensemble Methods}
We observe that previous ensemble methods underperform compared to the best candidate model in the Flores En$\rightarrow$Ro, Flores En$\rightarrow$De, and MBPP tasks. This underperformance stems from the inherent limitations of these methods, specifically:

\paragraph{Token-level ensemble is constrained by insufficient and inaccurate information.}
Firstly, since tokens, as units smaller than words, carry very limited information, it becomes extremely challenging to determine which option at the current step is most beneficial for the overall outcome.
Secondly, token-level ensemble methods rely on overlapping tokens as anchors to align the output distributions from heterogeneous LLMs into the same space, which introduces additional noise. As shown in Figures~\ref{fig:main_result} (d) and (f), the performance of EVA is significantly lower than that of the best candidate model and also falls well short of \textsc{SWeetSpan} with one-word span. Since tokens in non-English languages like German and Romanian are less familiar to candidate LLMs, the negative impact of noise outweighs the benefits of the ensemble approach.
In summary, token-level ensembles with insufficient and inaccurate information lead to suboptimal results at each generation step, hindering the achievement of globally optimal outcomes.

Our proposed \textsc{SweetSpan} avoids crossing word boundaries, offering a simple and effective way to eliminate the noise caused by different tokenizations across models. In addition, it ensures that the necessary information is captured through an appropriate span length, resulting in consistent performance improvements across tasks.

\paragraph{Sample-level ensemble is constrained by training data exposure.}
Sample-level ensemble methods rely on training an additional reward or fusion model to select or combine all candidate answers, which poses significant challenges in generalizing to unseen data distributions. 
For benchmarks in unfamiliar domains, such as code generation or non-English language translation tasks, their performance can degrade significantly.
As shown in Figure~\ref{fig:main_result} (b), LLM-Blender performs very poorly on the code generation task, indicating that its generative fusion model lacks the capability to effectively generate code.

In contrast, our proposed \textsc{SweetSpan} is a training-free method, thereby avoiding the generalization issues associated with training data exposure and offering superior generalization across tasks. 

\begin{table*}[!t]
\centering
\fontsize{10.0}{11.5} \selectfont
\resizebox{\textwidth}{!}{
\begin{tabular}{@{}lcccccc@{}}
\toprule
 & \multicolumn{3}{c}{\textbf{MBPP}}                       & \multicolumn{3}{c}{\textbf{GSM8K}} \ \\ \cmidrule(l){2-4} \cmidrule(l){5-7}
\ \textbf{System}         & \textbf{4 Good} & \textbf{3 Good \& 1 Bad}           & \textbf{2 Good \& 2 Bad}  &  \textbf{4 Good} & \textbf{3 Good \& 1 Bad}           & \textbf{2 Good \& 2 Bad}   \\    \midrule
\ LLaMA2-7B-Chat & \textcolor{lightgray}{17.93} & \textcolor{lightgray}{17.93} & \colorbox{red!20}{17.93} & \textcolor{lightgray}{25.02} & \textcolor{lightgray}{25.02} & \textcolor{lightgray}{25.02} \   \\
\ ChatGLM2-6B & \textcolor{lightgray}{16.60} & \colorbox{red!40}{16.60} & \colorbox{red!40}{16.60} & \colorbox{green!50}{30.63} & \colorbox{green!50}{30.63} & \colorbox{green!50}{30.63}  \      \\
\ Baichuan2-7B-Chat & \colorbox{green!50}{29.53} & \colorbox{green!50}{29.53} & \colorbox{green!50}{29.53} & \colorbox{green!30}{28.58}  & \colorbox{green!30}{28.58} & \textcolor{lightgray}{28.58}      \     \\
\ InternLM-7B-Chat & \colorbox{green!30}{27.20} & \colorbox{green!30}{27.20} & \textcolor{lightgray}{27.20} & \colorbox{green!70}{33.51} & \colorbox{green!70}{33.51} & \colorbox{green!70}{33.51}  \     \\
\ TigerBot-7B-Chat-V3  & \colorbox{green!70}{32.53} & \colorbox{green!70}{32.53} & \colorbox{green!70}{32.53} & \colorbox{green!15}{28.20} & \textcolor{lightgray}{28.20} & \textcolor{lightgray}{28.20}    \    \\
\ Vicuna-7B-V1.5 & \textcolor{lightgray}{-} & \textcolor{lightgray}{-} & \textcolor{lightgray}{-} & \textcolor{lightgray}{18.12}   & \textcolor{lightgray}{18.12} & \colorbox{red!20}{18.12}                 \    \\
\ ChineseAlpaca2-7B & \colorbox{green!15}{22.80} & \textcolor{lightgray}{22.80} & \textcolor{lightgray}{22.80} & \textcolor{lightgray}{10.08} & \colorbox{red!40}{10.08} & \colorbox{red!40}{10.08}        \    \\ \midrule
\ PairRanker~\citep{llm-blender-2023} & 26.87(\textit{-\ \ 5.66}) & 19.47(\textit{-13.06}) & 17.60(\textit{-14.93}) & 36.47(\textit{+\ 2.96}) & 33.28(\textit{-\ 0.23}) & 28.28(\textit{-\ 5.23}) \   \\
\ LLM-Blender~\citep{llm-blender-2023} & \ \ 5.33(\textit{-27.20}) & \ \ 5.13(\textit{-27.40}) & \ \ 4.00(\textit{-28.53}) & 34.27(\textit{+ 0.76}) & 29.04(\textit{-\ 4.47}) & 28.96(\textit{-\ 4.55}) \ \\ 
\ EVA~\citep{xu2024bridging} & 23.87(\textit{-\ \ 8.66}) & 28.99(\textit{-\ \ 3.54}) & 23.08(\textit{-\ \ 9.45}) & 41.32(\textit{+ 7.81}) &  37.15(\textit{+3.64}) & 26.08(\textit{-\ 7.43})  \   \\ \midrule
\ \textbf{SweetSpan} (\textit{ours}) & \textbf{38.20(\textit{+\ 5.67})} & \textbf{37.07(\textit{+\ 4.54})} & \textbf{35.53(\textit{+\ 3.00})}   & \textbf{46.32(\textit{+12.81})} & \textbf{39.20(\textit{+5.69})} & \textbf{36.32(\textit{+2.81})} \   \\ \bottomrule
\end{tabular}
}
\caption{\label{table:hard_result1} 
Main results on the code generation task (MBPP) and the arithmetic reasoning task (GSM8K) under the challenge setting. Best results are highlighted with bold and the models employed within the ensemble are distinguished by color according to their performance.
Vicuna is excluded from the MBPP task because we find that it cannot generate properly fomatted executable code.}
\end{table*}

\begin{table*}[!ht]
\centering
\fontsize{10.0}{11.5} \selectfont
\resizebox{\textwidth}{!}{
\begin{tabular}{@{}lcccccc@{}}
\toprule
 & \multicolumn{3}{c}{\textbf{NQ}}                       & \multicolumn{3}{c}{\textbf{Flores De$\rightarrow$En}} \ \\ \cmidrule(l){2-4} \cmidrule(l){5-7}
\ \textbf{System}         & \textbf{4 Good} & \textbf{3 Good \& 1 Bad}           & \textbf{2 Good \& 2 Bad}  &  \textbf{4 Good} & \textbf{3 Good \& 1 Bad}           & \textbf{2 Good \& 2 Bad}   \\    \midrule
\ LLaMA2-7B-Chat & \colorbox{green!70}{29.34} & \colorbox{green!70}{29.34} & \colorbox{green!70}{29.34} & \colorbox{green!50}{42.36} & \colorbox{green!50}{42.36} & \colorbox{green!50}{42.36} \   \\
\ ChatGLM2-6B & \textcolor{lightgray}{14.68} & \colorbox{red!40}{14.68} & \colorbox{red!40}{14.68} & \textcolor{lightgray}{34.70} & \textcolor{lightgray}{34.70} & \colorbox{red!20}{34.70}  \      \\
\ Baichuan2-7B-Chat & \colorbox{green!30}{24.40} & \colorbox{green!30}{24.40} & \textcolor{lightgray}{24.40} & \colorbox{green!30}{40.38}  & \colorbox{green!30}{40.38} & \textcolor{lightgray}{40.38}      \     \\
\ InternLM-7B-Chat & \textcolor{lightgray}{16.79} & \textcolor{lightgray}{16.79} & \colorbox{red!20}{16.79}  & \textcolor{lightgray}{33.18} & \colorbox{red!40}{33.18} & \colorbox{red!40}{33.18}  \     \\
\ TigerBot-7B-Chat-V3  & \colorbox{green!15}{23.49} & \textcolor{lightgray}{23.49} & \textcolor{lightgray}{23.49} & \textcolor{lightgray}{36.21} & \textcolor{lightgray}{36.21} & \textcolor{lightgray}{36.21}   \    \\
\ Vicuna-7B-V1.5 & \colorbox{green!50}{27.26} & \colorbox{green!50}{27.26} & \colorbox{green!50}{27.26} & \colorbox{green!70}{42.84} & \colorbox{green!70}{42.84} & \colorbox{green!70}{42.84}    \    \\
\ ChineseAlpaca2-7B & \textcolor{lightgray}{22.83} & \textcolor{lightgray}{22.83} & \textcolor{lightgray}{22.83} & \colorbox{green!15}{39.08} & \textcolor{lightgray}{39.08} & \textcolor{lightgray}{39.08}       \    \\ \midrule
\ PairRanker~\citep{llm-blender-2023} & 30.61(\textit{+1.27}) & 30.69(\textit{+1.35}) & 30.55(\textit{+1.21}) & 40.05(\textit{-\ 2.79}) & 36.82(\textit{-\ 6.02}) & 35.24(\textit{-\ 7.60}) \   \\
\ LLM-Blender~\citep{llm-blender-2023} & 31.99(\textit{+2.65}) & 31.52(\textit{+2.18}) & \textbf{30.89(\textit{+1.55})} & 44.11(\textit{+1.27}) & 43.90(\textit{+1.06}) & 43.42(\textit{+0.58}) \ \\ 
\ EVA~\citep{xu2024bridging} & 31.14(\textit{+1.80}) & 30.22(\textit{+0.88}) & 28.64(\textit{-\ 0.70}) & 43.34(\textit{+0.50}) &  43.69(\textit{+0.85}) &  42.44(\textit{-\ 0.40}) \   \\ \midrule
\ \textbf{SweetSpan} (\textit{ours}) & \textbf{33.38(\textit{+4.04})} & \textbf{32.55(\textit{+3.21})} & 30.30(\textit{+0.96})   & \textbf{44.78(\textit{+1.94})} & \textbf{44.82(\textit{+1.98})} & \textbf{43.76(\textit{+0.92})}  \   \\ \bottomrule
\end{tabular}
}
\caption{\label{table:hard_result2} 
Main results on the commonsense reasoning task (NQ) and the machine translation task (Flores De$\rightarrow$En) under the challenge setting. }
\end{table*}

\subsection{Challenging Setting: Ensembling Models with Substantial Performance Gaps}\label{sec:exp2}
Previous work focuses on ensembling models with similar performance~\cite{huang2024enabling}. 
However, in real-world scenarios, it is inevitable to encounter underperforming models in an ensemble when dealing with unknown tasks. To investigate this, we introduce a new challenge setting, ensembling models with significant performance gaps, to evaluate the noise resistance of ensemble methods.
Specifically, we perform experiments under two challenging settings\footnote{The scenario most representative of real-world situations involves randomly selecting four models for ensembling. Recognizing the impracticality of exhaustively evaluating every possible combination, we select models with the largest performance gaps to simulate the most challenging scenario among all potential situations.}: one combining the top three models with the worst model, and another combining the top two with the bottom two models in each task.
Experimental results on NQ, GSM8K, MBPP, and Flores-De$\rightarrow$En are shown in Table~\ref{table:hard_result1} and Table~\ref{table:hard_result2}.

\subsubsection{\textsc{SweetSpan} Demonstrates Robustness}
We observe that previous ensemble methods collapse on most tasks under this setting, while our proposed \textsc{SweetSpan} consistently delivers stable performance improvements over individual models and token-level ensemble methods across the four tasks, demonstrating its robustness and versatility.
Notably, \textsc{SweetSpan} also outperforms LLM-Blender, which utilizes an additional 3B-parameter fusion model, on three of four tasks, further demonstrating the effectiveness of our approach.
We attribute the robustness of our method to the effective filtering strategy that excludes outliers with abnormal perplexity, thereby preventing unfaithful evaluations from impacting the ensemble results.

\subsection{Ablation Study on Filtering Strategy}
To validate the effectiveness of the filtering strategy in \textsc{SweetSpan}, we conduct ablation studies in both standard and challenging settings. Experimental results on MBPP, GSM8K, NQ, and Flores-De$\rightarrow$En are presented in Table~\ref{tab:ablation}.

Overall, the filtering strategy in \textsc{SweetSpan} consistently enhances performance across tasks. As the settings become increasingly challenging, the impact of the filtering strategy becomes more pronounced. 
We observe that the filtering strategy yields modest improvements in tasks with relatively stable performance, such as machine translation. However, for tasks significantly affected by underperforming models, such as MBPP, the strategy results in substantial gains. Notably, when the ensemble includes the two lowest-performing models, it achieves a 10.93\% increase in Pass@1 on MBPP compared to using no filtering, highlighting its critical role in maintaining \textsc{SweetSpan}'s robustness in challenging scenarios. 

\begin{table}[!t]
\centering
\resizebox{\linewidth}{!}{
\begin{tabular}{@{}lcccc@{}}
\toprule
\ \textbf{Methods}         & \textbf{MBPP}  & \textbf{GSM8K} & \textbf{NQ} & \textbf{Flores De$\rightarrow$En}   \\ \midrule
 \multicolumn{5}{c}{\textit{4 Good}}  \\ \hdashline
\ \textsc{SweetSpan}       & 38.20 & 46.32 & 33.38 & 44.78 \\
\ w/o Filter               & 36.80 & 46.02 & 31.39 & 44.78 \\ \midrule
 \multicolumn{5}{c}{\textit{3 Good \& 1 Bad}}  \\ \hdashline
\ \textsc{SweetSpan}       & 37.07 & 39.20 & 32.55 & 44.82 \\
\ w/o Filter               & 34.00 & 38.51 & 30.97 & 44.79 \\ \midrule
 \multicolumn{5}{c}{\textit{2 Good \& 2 Bad}}  \\ \hdashline
\ \textsc{SweetSpan}       & 35.53 & 36.32 & 30.30 & 43.76 \\
\ w/o Filter               & 24.60 & 35.33 & 25.37 & 43.26 \\
\bottomrule
\end{tabular}
}
\caption{\label{tab:ablation} 
Ablation study on filtering strategy}
\end{table}

\section{Related Work}
Ensemble learning is a widely adopted technique to enhance performance on specific tasks and ensure robust generalization by combining multiple complementary systems~\citep{zhou2017neural,liu2018comparable, lu2024merge}. 
Based on the granularity of the ensemble, existing ensemble methods can be categorized into two categories: sample-level ensemble and token-level ensemble.

\noindent\textbf{Sample-level Ensemble}
Sample-level ensemble can be further categorized into selection-based ensemble and fusion-based ensemble.
Selection-based ensemble methods select the best LLM for specific examples before inference or select the best output from multiple outputs after inference.
\citet{shnitzer2023large} uses benchmark datasets to train a router model that selects the best LLM out of a collection of models for a given task. 
\citet{lu2023routing} introduce ZOOTER, a system that uses a reward model to score query-output pairs, then trains a router via knowledge distillation to select the optimal LLM based on input queries.
FrugalGPT~\citep{chen2023frugalgpt} sequentially calls LLMs until a scoring model deems the output acceptable, efficiently utilizing multiple LLMs.
Such methods are limited by the output quality of the candidate models and are unable to generate outputs superior to those of existing models.

Unlike selection-based methods, fusion-based ensembles bypass the limitations of complete outputs, often yielding better results.
\citet{llm-blender-2023} select the top K outputs with a pair ranker, then use a fusion model to combine them. However, this approach relies on the generative capacity of the fusion model, which is limited by its exposure to training data. Moreover, the use of a fusion model considerably increases training and inference costs. For instance, \citet{llm-blender-2023} utilizes a 3B-sized fusion model.

Our proposed \textsc{SweetSpan} is a training-free approach, avoiding the generalization issues tied to training data exposure and providing superior generalization across tasks.

\noindent\textbf{Token-level Ensemble}
Token-level ensemble methods combine the output distribution of candidate models at each generation step.
Several studies fuse LLMs with specialized models that share the same vocabulary to enhance the specific capabilities of LLMs. 
\citet{li2024purifying} combine untrusted LLMs with a benign smaller LLM to mitigate issues such as copyright infringement, data poisoning, and privacy violations. \citet{hoang2024fly} enhance translation performance by ensembling a machine translation model with an LLM.

However, most open-source LLMs are heterogeneous and have different vocabularies, hindering direct ensembling.
\citet{fu2023specializing} and \citet{wan2024knowledge} employ exact match constraints and minimum edit distance strategy, respectively, to align vocabularies.
\citet{xu2024bridging}, \citet{huang2024enabling} and \citet{yu2024breaking} utilize overlapping tokens as anchors to map the output distributions of heterogeneous LLMs into a unified space.
Specifically, \citet{xu2024bridging} propose to directly learn the projection matrices between different vocabularies using the anchors as bridges, while \citet{huang2024enabling} and \citet{yu2024breaking} calculate the relative representations from anchors to different vocabularies, thereby indirectly achieving the vocabulary projection.
The process of vocabulary alignment introduces noise, which constrains the effectiveness of these methods.

Our proposed \textsc{SweetSpan} does not cross word boundaries, eliminating the noise caused by varying tokenizers.

\section{Conclusion}
In this paper, we propose a span-level ensemble method, \textsc{SweetSpan}, which balances the flexibility needed for real-time adjustments and the information required for accurate ensemble decisions at each step.
Our method has no limitations regarding vocabulary, model architecture, or task, and can be directly applied to any LLM ensemble without additional parameter training, making it a simple and versatile approach.
Experimental results in both standard and challenging settings demonstrate the superiority of our approach, which stably and significantly improves overall performance across various natural language processing tasks.

\section*{Limitations}
\paragraph{Efficiency}
Due to the inherent nature of ensembling, our approach, like previous ensemble methods, requires performing inference N times when ensembling N models. Additionally, we need to compute perplexity through forward propagation. However, we argue that the inferences for generating candidate spans and computing perplexity can be executed in parallel, respectively, as they are completely independent processes. Notably, our method demonstrates superior efficiency compared to the token-level approach, and the time overhead decreases as the span length increases. Please refer to Appendix~\ref{sec:efficiency} for detailed efficiency analysis.

\paragraph{LLM Evaluation} 
Considering the expenses, we do not use human or GPT-4 evaluation. 
Human or GPT-4 evaluation could provide us with more reliable and comprehensive results. However, due to the number of models and datasets in our experiments, we cannot afford large-scale human evaluation.


\bibliography{custom}

\begin{thebibliography}{27}
\providecommand{\natexlab}[1]{#1}

\bibitem[{Austin et~al.(2021)Austin, Odena, Nye, Bosma, Michalewski, Dohan, Jiang, Cai, Terry, Le et~al.}]{austin2021program}
Jacob Austin, Augustus Odena, Maxwell Nye, Maarten Bosma, Henryk Michalewski, David Dohan, Ellen Jiang, Carrie Cai, Michael Terry, Quoc Le, et~al. 2021.
\newblock Program synthesis with large language models.
\newblock \emph{arXiv preprint arXiv:2108.07732}.

\bibitem[{Baichuan(2023)}]{baichuan2023baichuan2}
Baichuan. 2023.
\newblock \href {https://arxiv.org/abs/2309.10305} {Baichuan 2: Open large-scale language models}.
\newblock \emph{arXiv preprint arXiv:2309.10305}.

\bibitem[{Chen et~al.(2023{\natexlab{a}})Chen, Zaharia, and Zou}]{chen2023frugalgpt}
Lingjiao Chen, Matei Zaharia, and James Zou. 2023{\natexlab{a}}.
\newblock Frugalgpt: How to use large language models while reducing cost and improving performance.
\newblock \emph{arXiv preprint arXiv:2305.05176}.

\bibitem[{Chen et~al.(2023{\natexlab{b}})Chen, Cai, Wu, Li, Xin, and Fu}]{chen2023tigerbot}
Ye~Chen, Wei Cai, Liangmin Wu, Xiaowei Li, Zhanxuan Xin, and Cong Fu. 2023{\natexlab{b}}.
\newblock Tigerbot: An open multilingual multitask llm.
\newblock \emph{arXiv preprint arXiv:2312.08688}.

\bibitem[{Chiang et~al.(2023)Chiang, Li, Lin, Sheng, Wu, Zhang, Zheng, Zhuang, Zhuang, Gonzalez, Stoica, and Xing}]{vicuna2023}
Wei-Lin Chiang, Zhuohan Li, Zi~Lin, Ying Sheng, Zhanghao Wu, Hao Zhang, Lianmin Zheng, Siyuan Zhuang, Yonghao Zhuang, Joseph~E. Gonzalez, Ion Stoica, and Eric~P. Xing. 2023.
\newblock \href {https://lmsys.org/blog/2023-03-30-vicuna/} {Vicuna: An open-source chatbot impressing gpt-4 with 90\%* chatgpt quality}.

\bibitem[{Cobbe et~al.(2021)Cobbe, Kosaraju, Bavarian, Chen, Jun, Kaiser, Plappert, Tworek, Hilton, Nakano, Hesse, and Schulman}]{cobbe2021gsm8k}
Karl Cobbe, Vineet Kosaraju, Mohammad Bavarian, Mark Chen, Heewoo Jun, Lukasz Kaiser, Matthias Plappert, Jerry Tworek, Jacob Hilton, Reiichiro Nakano, Christopher Hesse, and John Schulman. 2021.
\newblock Training verifiers to solve math word problems.
\newblock \emph{arXiv preprint arXiv:2110.14168}.

\bibitem[{Cui et~al.(2023)Cui, Yang, and Yao}]{Chinese-LLaMA-Alpaca}
Yiming Cui, Ziqing Yang, and Xin Yao. 2023.
\newblock \href {https://arxiv.org/abs/2304.08177} {Efficient and effective text encoding for chinese llama and alpaca}.
\newblock \emph{arXiv preprint arXiv:2304.08177}.

\bibitem[{Fu et~al.(2023)Fu, Peng, Ou, Sabharwal, and Khot}]{fu2023specializing}
Yao Fu, Hao Peng, Litu Ou, Ashish Sabharwal, and Tushar Khot. 2023.
\newblock Specializing smaller language models towards multi-step reasoning.
\newblock In \emph{International Conference on Machine Learning}, pages 10421--10430. PMLR.

\bibitem[{Goyal et~al.(2022)Goyal, Gao, Chaudhary, Chen, Wenzek, Ju, Krishnan, Ranzato, Guzm{\'a}n, and Fan}]{flores101}
Naman Goyal, Cynthia Gao, Vishrav Chaudhary, Peng-Jen Chen, Guillaume Wenzek, Da~Ju, Sanjana Krishnan, Marc’Aurelio Ranzato, Francisco Guzm{\'a}n, and Angela Fan. 2022.
\newblock The flores-101 evaluation benchmark for low-resource and multilingual machine translation.
\newblock \emph{Transactions of the Association for Computational Linguistics}, 10:522--538.

\bibitem[{Hoang et~al.(2024)Hoang, Khayrallah, and Junczys-Dowmunt}]{hoang2024fly}
Hieu Hoang, Huda Khayrallah, and Marcin Junczys-Dowmunt. 2024.
\newblock On-the-fly fusion of large language models and machine translation.
\newblock In \emph{Findings of the Association for Computational Linguistics: NAACL 2024}, pages 520--532.

\bibitem[{Huang et~al.(2024)Huang, Feng, Li, Xiang, Wang, Qin, and Liu}]{huang2024enabling}
Yichong Huang, Xiaocheng Feng, Baohang Li, Yang Xiang, Hui Wang, Bing Qin, and Ting Liu. 2024.
\newblock Enabling ensemble learning for heterogeneous large language models with deep parallel collaboration.
\newblock \emph{arXiv preprint arXiv:2404.12715}.

\bibitem[{Jiang et~al.(2023)Jiang, Ren, and Lin}]{llm-blender-2023}
Dongfu Jiang, Xiang Ren, and Bill~Yuchen Lin. 2023.
\newblock Llm-blender: Ensembling large language models with pairwise comparison and generative fusion.
\newblock In \emph{Proceedings of the 61th Annual Meeting of the Association for Computational Linguistics (ACL 2023)}.

\bibitem[{Kwiatkowski et~al.(2019)Kwiatkowski, Palomaki, Redfield, Collins, Parikh, Alberti, Epstein, Polosukhin, Devlin, Lee et~al.}]{kwiatkowski2019natural}
Tom Kwiatkowski, Jennimaria Palomaki, Olivia Redfield, Michael Collins, Ankur Parikh, Chris Alberti, Danielle Epstein, Illia Polosukhin, Jacob Devlin, Kenton Lee, et~al. 2019.
\newblock Natural questions: a benchmark for question answering research.
\newblock \emph{Transactions of the Association for Computational Linguistics}, 7:453--466.

\bibitem[{Li et~al.(2024)Li, Liu, Pang, Du, Guo, Liu, and Lin}]{li2024purifying}
Tianlin Li, Qian Liu, Tianyu Pang, Chao Du, Qing Guo, Yang Liu, and Min Lin. 2024.
\newblock Purifying large language models by ensembling a small language model.
\newblock \emph{arXiv preprint arXiv:2402.14845}.

\bibitem[{Liu et~al.(2018)Liu, Zhou, Wang, Zhao, Zhang, and Zong}]{liu2018comparable}
Yuchen Liu, Long Zhou, Yining Wang, Yang Zhao, Jiajun Zhang, and Chengqing Zong. 2018.
\newblock A comparable study on model averaging, ensembling and reranking in nmt.
\newblock In \emph{Natural Language Processing and Chinese Computing: 7th CCF International Conference, NLPCC 2018, Hohhot, China, August 26--30, 2018, Proceedings, Part II 7}, pages 299--308. Springer.

\bibitem[{Lu et~al.(2024)Lu, Pang, Xiao, Zhu, Xia, and Zhang}]{lu2024merge}
Jinliang Lu, Ziliang Pang, Min Xiao, Yaochen Zhu, Rui Xia, and Jiajun Zhang. 2024.
\newblock Merge, ensemble, and cooperate! a survey on collaborative strategies in the era of large language models.
\newblock \emph{arXiv preprint arXiv:2407.06089}.

\bibitem[{Lu et~al.(2023)Lu, Yuan, Lin, Lin, Yuan, Zhou, and Zhou}]{lu2023routing}
Keming Lu, Hongyi Yuan, Runji Lin, Junyang Lin, Zheng Yuan, Chang Zhou, and Jingren Zhou. 2023.
\newblock Routing to the expert: Efficient reward-guided ensemble of large language models.
\newblock \emph{arXiv preprint arXiv:2311.08692}.

\bibitem[{Post(2018)}]{post-2018-call}
Matt Post. 2018.
\newblock \href {https://doi.org/10.18653/v1/W18-6319} {A call for clarity in reporting {BLEU} scores}.
\newblock In \emph{Proceedings of the Third Conference on Machine Translation: Research Papers}, pages 186--191, Brussels, Belgium. Association for Computational Linguistics.

\bibitem[{Shnitzer et~al.(2023)Shnitzer, Ou, Silva, Soule, Sun, Solomon, Thompson, and Yurochkin}]{shnitzer2023large}
Tal Shnitzer, Anthony Ou, M{\'\i}rian Silva, Kate Soule, Yuekai Sun, Justin Solomon, Neil Thompson, and Mikhail Yurochkin. 2023.
\newblock Large language model routing with benchmark datasets.
\newblock \emph{arXiv preprint arXiv:2309.15789}.

\bibitem[{Team(2023)}]{2023internlm}
InternLM Team. 2023.
\newblock Internlm: A multilingual language model with progressively enhanced capabilities.
\newblock \url{https://github.com/InternLM/InternLM}.

\bibitem[{Touvron et~al.(2023)Touvron, Martin, Stone, Albert, Almahairi, Babaei, Bashlykov, Batra, Bhargava, Bhosale et~al.}]{touvron2023llama}
Hugo Touvron, Louis Martin, Kevin Stone, Peter Albert, Amjad Almahairi, Yasmine Babaei, Nikolay Bashlykov, Soumya Batra, Prajjwal Bhargava, Shruti Bhosale, et~al. 2023.
\newblock Llama 2: Open foundation and fine-tuned chat models.
\newblock \emph{arXiv preprint arXiv:2307.09288}.

\bibitem[{Uzan et~al.(2024)Uzan, Schmidt, Tanner, and Pinter}]{uzan2024greed}
Omri Uzan, Craig~W Schmidt, Chris Tanner, and Yuval Pinter. 2024.
\newblock Greed is all you need: An evaluation of tokenizer inference methods.
\newblock \emph{arXiv preprint arXiv:2403.01289}.

\bibitem[{Wan et~al.(2024)Wan, Huang, Cai, Quan, Bi, and Shi}]{wan2024knowledge}
Fanqi Wan, Xinting Huang, Deng Cai, Xiaojun Quan, Wei Bi, and Shuming Shi. 2024.
\newblock Knowledge fusion of large language models.
\newblock \emph{arXiv preprint arXiv:2401.10491}.

\bibitem[{Xu et~al.(2024)Xu, Lu, and Zhang}]{xu2024bridging}
Yangyifan Xu, Jinliang Lu, and Jiajun Zhang. 2024.
\newblock Bridging the gap between different vocabularies for llm ensemble.
\newblock In \emph{Proceedings of the 2024 Conference of the North American Chapter of the Association for Computational Linguistics: Human Language Technologies (Volume 1: Long Papers)}, pages 7133--7145.

\bibitem[{Yu et~al.(2024)Yu, Kuo, Ye, Chang, and Li}]{yu2024breaking}
Yao-Ching Yu, Chun-Chih Kuo, Ziqi Ye, Yu-Cheng Chang, and Yueh-Se Li. 2024.
\newblock Breaking the ceiling of the llm community by treating token generation as a classification for ensembling.
\newblock \emph{arXiv preprint arXiv:2406.12585}.

\bibitem[{Zeng et~al.(2022)Zeng, Liu, Du, Wang, Lai, Ding, Yang, Xu, Zheng, Xia et~al.}]{zeng2022glm}
Aohan Zeng, Xiao Liu, Zhengxiao Du, Zihan Wang, Hanyu Lai, Ming Ding, Zhuoyi Yang, Yifan Xu, Wendi Zheng, Xiao Xia, et~al. 2022.
\newblock Glm-130b: An open bilingual pre-trained model.
\newblock \emph{arXiv preprint arXiv:2210.02414}.

\bibitem[{Zhou et~al.(2017)Zhou, Hu, Zhang, and Zong}]{zhou2017neural}
Long Zhou, Wenpeng Hu, Jiajun Zhang, and Chengqing Zong. 2017.
\newblock Neural system combination for machine translation.
\newblock In \emph{Proceedings of the 55th Annual Meeting of the Association for Computational Linguistics (Volume 2: Short Papers)}, pages 378--384.

\end{thebibliography}

\appendix

\section{Effect of Model Filtering Intensity}
\label{sec:lambda}
In Section~\ref{sec:method2}, we introduced the hyperparameter $\lambda$ to control the filtering threshold for the evaluation scores.
A smaller $\lambda$ results in more aggressive filtering. For example, when $\lambda=0$, each span has its highest and lowest scores removed. In contrast, a larger $\lambda$ results in a more lenient filtering, with filtering bypassed as $\lambda$ becomes very large.
We conduct experiments on the NQ and MBPP tasks to evaluate the effect of different $\lambda$ values on the ensemble results. Since the optimal $\lambda$ varies across tasks, we select $\lambda=10$ as a balanced option for all experiments and observe consistent performance across various tasks.

\begin{table}[!h]
\centering
\begin{tabular}{@{}lcc@{}}
\toprule
\ $\lambda$ & NQ & MBPP \  \\ \midrule
\ 0         & 33.05 & 40.93\  \\
\ 10       & 33.38 & 38.20\  \\
\ 20       & 33.30 & 37.53\  \\
\ 40       & 33.38 & 37.80\  \\ \bottomrule
\end{tabular}
\caption{\label{table:lambda} 
Effect of Model Filtering Intensity.}
\end{table}
\section{Efficiency Analysis}
\label{sec:efficiency}
The time overhead of ensemble methods during the generation process, such as span-level ensemble and token-level ensemble, is related to the length of the generated text. The longer the generated text, the greater the time overhead. As shown in Table~\ref{table:efficiency}, we compare the efficiency of \textsc{SweetSpan} with the token-level ensemble method EVA by calculating the average extra time required per token. 
Our method shows superiority in efficiency compared to the token-level approach, and the time overhead decreases as the span length increases.

\begin{table}[!h]
\centering
\begin{tabular}{@{}lc@{}}
\toprule
\ Method & Average Extra Time Per Token  \\ \midrule
\ EVA          & 0.0777s  \\
\ Span-1       & 0.0512s  \\
\ Span-2       & 0.0224s  \\
\ Span-4       & 0.0128s  \\
\ Span-8       & 0.0079s  \\ 
\ Span-16      & 0.0053s  \\
\ Span-32      & 0.0077s \\\bottomrule
\end{tabular}
\caption{\label{table:efficiency} 
Efficiency Comparison of Token-level and Span-level Ensembles.}
\end{table}

\section{Main Results under the Standard Setting}
The main results on various language generation tasks under the standard setting are presented in Table~\ref{table:main_result1} and Table~\ref{table:main_result2}.

\label{sec:main_result}
\begin{table*}[!]
\centering
\begin{tabular}{@{}lccc@{}}
\toprule
\ \textbf{System}     & \textbf{NQ} & \textbf{GSM8K} & \textbf{MBPP} \\ \midrule
\ LLaMA2-7B-Chat      & {\ul 29.34} & 25.02          & 17.93         \\
\ ChatGLM2-6B         & 14.68       & {\ul 30.63}   & 16.60         \\
\ Baichuan2-7B-Chat   & {\ul 24.40} & {\ul 28.58}   & {\ul 29.53}   \\
\ InternLM-7B-Chat    & 16.79       & {\ul 33.51}    & {\ul 27.20}   \\
\ TigerBot-7B-Chat-V3 & {\ul 23.49} & {\ul 28.20}   & {\ul 32.53}   \\
\ Vicuna-7B-V1.5      & {\ul 27.26} & 18.12          & -            \\
\ ChineseAlpaca2-7B   & 22.83       & 10.08         & {\ul 22.80}  \\ \midrule
\ PairRanker          & 30.61(+\textit{1.27})                    & 36.47(+\textit{2.96})                       & 26.87(-\textit{5.66})                      \\
\ LLM-Blender         & 31.99(+\textit{2.65})                    & 34.27(+\textit{0.76})                       & 5.33(-\textit{27.20})                      \\
\ EVA                 & 31.14(+\textit{1.80})                    & 41.32(+\textit{7.81})                       & 23.87(-\textit{8.66})                      \\ \midrule
\ Span-1              & \textbf{33.38(+\textit{4.04})}           & 45.19(+\textit{11.68})                      & \textbf{38.20(+\textit{5.67})}              \\
\ Span-2              & 32.74(+\textit{3.40})                    & 44.73(+\textit{11.22})                      & 35.60(+\textit{3.07})                       \\
\ Span-4              & 33.10(+\textit{3.76})                    & 45.72(+\textit{12.21})                      & 34.67(+\textit{2.14})                      \\
\ Span-8              & 32.96(+\textit{3.62})                    & 46.32(+\textit{12.81})                      & 33.40(+\textit{0.87})                       \\
\ Span-16             & 32.05(+\textit{2.71})                    & \textbf{46.32(+\textit{12.81})}             & 34.40(+\textit{1.87})                       \\
\ Span-32             & 32.11(+\textit{2.77})                    & 45.41(+\textit{11.90})                       & 34.40(+\textit{1.87})                       \\ \bottomrule
\end{tabular}
\caption{\label{table:main_result1} 
Main results of NQ (measured by Exact Match), GSM8K (measured by Accuracy) and MBPP (measured by Pass@1). Best results are highlighted with bold and the model employed within the ensemble is underlined for distinction.
}
\end{table*}

\begin{table*}[!]
\centering

\begin{tabular}{@{}lcccc@{}}
\toprule
\ \textbf{System}     & \textbf{Flores-En$\rightarrow$De} & \textbf{Flores-De$\rightarrow$En} & \textbf{Flores-En$\rightarrow$Ro} & \textbf{Flores-Ro$\rightarrow$En} \\ \midrule
\ LLaMA2-7B-Chat      & {\ul 27.37}           & {\ul 42.36}           & {\ul 26.11}           & {\ul 40.64}           \\
\ ChatGLM2-6B         & 13.20                 & 34.70                 & 12.44                 & 31.30                 \\
\ Baichuan2-7B-Chat   & {\ul 25.96}           & {\ul 40.38}           & {\ul 24.58}           & {\ul 39.41}           \\
\ InternLM-7B-Chat    & 9.20                  & 33.18                 & 10.03                 & 30.23                 \\
\ TigerBot-7B-Chat-V3 & 20.78                 & 36.21                 & {\ul 13.07}           & 33.32                 \\
\ Vicuna-7B-V1.5      & {\ul 31.60}           & {\ul 42.84}           & {\ul 29.35}           & {\ul 41.11}           \\
\ ChineseAlpaca2-7B   & {\ul 21.83}           & {\ul 39.08}           & 6.96                  & {\ul 34.35}           \\ \midrule
\ PairRanker          & 26.71(-\textit{4.89})          & 40.05(-\textit{2.79})          & 22.29(-\textit{7.06})          & 37.45(-\textit{3.66})          \\
\ LLM-Blender         & 29.04(-\textit{2.56})          & 44.11(+\textit{1.27})          & 26.01(-\textit{3.34})          & 42.28(+\textit{1.17})          \\
\ EVA                 & 20.15(-\textit{11.45})         & 43.34(+\textit{0.50})          & 22.60(-\textit{6.75})          & 41.52(+\textit{0.41})          \\ \midrule
\ Span-1              & 32.85(+\textit{1.25})          & 43.93(+\textit{1.09})          & 30.48(+\textit{1.13})          & 42.75(+\textit{1.64})          \\
\ Span-2              & 33.15(+\textit{1.55})          & 44.20(+\textit{1.36})          & \textbf{31.65(+\textit{2.30})} & 42.93(+\textit{1.82})          \\
\ Span-4              & \textbf{33.78(+\textit{2.18})} & 44.60(+\textit{1.76})          & 30.97(+\textit{1.62})          & 42.95(+\textit{1.84})          \\
\ Span-8              & 33.37(+\textit{1.77})          & \textbf{44.78(+\textit{1.94})} & 30.29(+\textit{0.94})          & \textbf{42.99(+\textit{1.88})} \\
\ Span-16             & 32.89(+\textit{1.29})          & 44.62(+\textit{1.78})          & 30.22(+\textit{0.87})          & 42.65(+\textit{1.54})          \\
\ Span-32             & 32.89(+\textit{1.29})          & 44.62(+\textit{1.78})          & 30.22(+\textit{0.87})          & 42.62(+\textit{1.51})          \\ \bottomrule
\end{tabular}
\caption{\label{table:main_result2} 
Main results of machine translation tasks (measured by BLEU). Best results are highlighted in bold and the model employed within the ensemble is underlined for distinction.
}
\end{table*}

\end{document}